\pdfoutput=1
\documentclass[11pt,a4paper,table]{article}
\usepackage[hyperref]{naaclhlt2019}
\usepackage{times}
\usepackage{latexsym}

\usepackage{url}

\aclfinalcopy 
\def\aclpaperid{287} 

\usepackage[T1]{fontenc}
\usepackage{amsmath}
\usepackage{enumitem}
\usepackage{multirow}
\usepackage{tikz}
\usepackage{tikz-dependency}
\usepackage[warn]{textcomp}
\usepackage{caption}
\usepackage{subcaption}
\usepackage{multirow}
\usepackage{etoolbox}
\usepackage{xr}
\usepackage{xfrac}
\usepackage{pgf}
\usepackage{collcell}
\usepackage{booktabs}

\makeatletter
\renewcommand{\@BIBLABEL}{\@emptybiblabel}
\newcommand{\@emptybiblabel}[1]{}
\makeatother

\newcommand{\com}[1]{}

\usetikzlibrary{shapes,shapes.misc}

\newcommand{\ApplyGradient}[1]{%
  \pgfmathsetmacro{\PercentColor}{(#1-0)/63.88}%
  \pgfmathsetmacro{\PercentInverse}{ifthenelse(\PercentColor > 70, 0, 100)}%
  \edef\x{\noexpand\cellcolor{red!\PercentColor}}\x\textcolor{black!\PercentInverse}{#1}%
}
\newcolumntype{R}{>{\collectcell\ApplyGradient}{c}<{\endcollectcell}}

\title{Content Differences in Syntactic and Semantic Representations}

\author{Daniel Hershcovich$^{1,2}$ \\
  \\\And
  Omri Abend$^2$ \\
  $^1$The Edmond and Lily Safra Center for Brain Sciences \\
  $^2$School of Computer Science and Engineering \\
  Hebrew University of Jerusalem \\
  \texttt{\{danielh,oabend,arir\}@cs.huji.ac.il}
  \\\And
  Ari Rappoport$^2$
}

\date{}

\begin{document}

\maketitle

\begin{abstract}
  Syntactic analysis plays an important role in semantic parsing,
  but the nature of this role remains a topic of ongoing debate.
  The debate has been constrained by the scarcity of empirical comparative studies between syntactic and semantic schemes,
  which hinders the development of parsing methods informed by the details of target schemes and constructions.
  We target this gap, and take Universal Dependencies (UD) and UCCA as a test case.
  After abstracting away from differences of convention or formalism,
  we find that most content divergences can be ascribed to: 
  (1) UCCA's distinction between a Scene and a non-Scene;
  (2) UCCA's distinction between primary relations, secondary ones and participants;
  (3) different treatment of multi-word expressions, and
  (4) different treatment of inter-clause linkage.
  We further discuss the long tail of cases where the two schemes take markedly
  different approaches.
  Finally, we show that the proposed comparison methodology can be used
  for fine-grained evaluation of UCCA parsing,
  highlighting both challenges and potential sources for improvement.
  The substantial differences between the schemes suggest that
  semantic parsers are likely to benefit downstream text understanding applications
  beyond their syntactic counterparts.
\end{abstract}

\section{Introduction}\label{sec:introduction}
  
  Semantic representations hold promise due to
  their ability to transparently reflect distinctions relevant for text understanding
  applications. For example, syntactic representations
  are usually sensitive to distinctions based on POS (part of speech), such as between compounds
  and possessives. Semantic schemes are  less likely to make
  this distinction since a possessive can often be paraphrased as a compound
  and vice versa (e.g., ``US president''/``president of the US''),
  but may distinguish different senses of possessives (e.g., ``some of the presidents'' and ``inauguration of the presidents'').

  Nevertheless, little empirical study has been done on what distinguishes semantic schemes from
  syntactic ones, which are still in many cases the backbone of text understanding systems. 
  Such studies are essential for 
  (1) determining whether and to what extent semantic methods should be adopted for text understanding applications;
  (2) defining better inductive biases for semantic parsers, and allowing better use of information encoded in syntax;
  (3) pointing at semantic distinctions unlikely to be resolved by syntax.

  The importance of such an empirical study is emphasized by the ongoing discussion as to what role syntax should
  play in semantic parsing, if any \cite{swayamdipta2018syntactic,strubell2018linguistically,P18-1192,C18-1233}.
  See \S\ref{sec:related_work}.

  This paper aims to address this gap,
  focusing on {\it content} differences.
  As a test case, we compare relatively similar schemes (\S\ref{sec:representations}):
  the syntactic Universal Dependencies \cite[UD; ][]{nivre2016universal},
  and the semantic Universal Conceptual Cognitive Annotation \cite[UCCA; ][]{abend2013universal}.
  
  We UCCA-annotate
  the entire web reviews section of the UD EWT corpus
  (\S\ref{sec:shared}),
  and develop a converter to assimilate UD and UCCA,
  which use formally different graphs
  (\S\ref{sec:methodology}).
  We then align their nodes, and identify which UCCA categories match which UD relations,
  and which are unmatched.

  Most content differences are due to (\S\ref{sec:analysis}):
  \begin{enumerate}[itemsep=0.159mm,leftmargin=5.95mm]
      \item UCCA's distinction between words and phrases that evoke Scenes (events) and ones that do not.
        For example, eventive and non-eventive nouns are treated differently in UCCA, but similarly in UD.
      \item UCCA's distinction between primary relations, secondary relations
        and Participants, in contrast to UD's core/non-core distinction.
      \item Different treatment of multi-word expressions (MWEs),
        where UCCA has a stronger tendency to explicitly mark them.
      \item UCCA's conflation of several syntactic realizations of inter-clause linkage,
        and disambiguation of other cases that UD treats similarly.
   \end{enumerate}

  We show that the differences between the schemes are substantial, and suggest that
  UCCA parsing in particular and semantic parsing in general are likely to benefit
  downstream text understanding applications.
  For example, only 72.9\% of UCCA Participants are UD syntactic arguments,
  i.e., many semantic participants cannot be recovered from UD.\footnote{This excludes cases of shared 
  argumenthood, which are partially covered by \textit{enhanced UD}. See \S\ref{sec:conversion}.}
  Our findings are relevant to other semantic representations, given their 
  significant overlap in content \cite{abend2017state}.
      
  A methodology for comparing syntactic and semantic treebanks can also support fine-grained error 
  analysis of semantic parsers, as illustrated by \citet{szubert2018structured} 
  for AMR \citep{banarescu2013abstract}.
  To demonstrate the utility of our comparison methodology,
  we perform fine-grained error analysis on UCCA parsing,
  according to UD relations (\S\ref{sec:fine_grained}).
  Results highlight challenges for current parsing technology,
  and expose cases where UCCA parsers may benefit from modeling syntactic structure more directly.\footnote{Our conversion and analysis code is public available
  at \url{https://github.com/danielhers/synsem}.}


\section{Representations}\label{sec:representations}

  The conceptual and formal similarity between UD and UCCA can be traced back
  to their shared design principles:
  both are designed to be applicable across languages and domains, 
  to enable rapid annotation and to support text understanding
  applications. This section provides a brief introduction to each of the schemes, whereas
  the next sections discuss their content in further
  detail.\footnote{See Supplementary Material for a definition of each category in both schemes,
  and their abbreviations.}

\paragraph{UCCA}\label{sec:ucca}
  is a semantic annotation scheme rooted in typological 
  and cognitive linguistic theory.
  It aims to represent the main semantic phenomena in text, abstracting away from syntactic forms.
  Shown to be preserved remarkably well across translations \citep{sulem2015conceptual}, it has been applied to
  improve text simplification \citep{sulem2018simple},
  and text-to-text generation evaluation \citep{birch2016hume,choshen2018usim,sulem2018samsa}.
  
  \begin{table}
  \small
  \centering
  \begin{tabular}{lc}
    Participant & A \\
    Center & C \\
    Adverbial & D \\
    Elaborator & E \\
    Function & F \\
    Ground & G \\
    Parallel Scene & H
 \end{tabular}
 \hfill
 \vrule
 \hfill
 \begin{tabular}{lc}
    Linker & L \\
    Connector & N \\
    Process & P \\
    Quantifier & Q \\
    Relator & R \\
    State & S \\
    Time & T
 \end{tabular}
 \caption{Legend of UCCA categories (edge labels).\label{fig:legend}}
 \end{table}

  Formally, UCCA structures are directed acyclic graphs (DAGs) whose nodes (or {\it units}) correspond either to words,
  or to elements viewed as a single entity according to some semantic or cognitive consideration.
  Edges are labeled, indicating the role of a child in the relation the parent represents.
  Figure~\ref{fig:legend} shows a legend of UCCA abbreviations.
  A {\it Scene} is UCCA's notion of an event or a frame, and is a description of a movement, an action or a state which persists in time. 
  Every Scene contains one primary relation, which can be either a Process or a State. 
  Scenes may contain any number of Participants, a category which also includes abstract participants and locations.
  They may also contain temporal relations (Time), and secondary relations (Adverbials), 
  which cover semantic distinctions such as manner, modality and aspect.\footnote{Despite the
  similar terminology, UCCA Adverbials are not necessarily adverbs syntactically.}

  Scenes may be \textit{linked} to one another in several ways.
  First, a Scene can provide information about some entity,
  in which case it is marked as an Elaborator.
  This often occurs in the case of participles or relative clauses.
  For example, ``(child) who went to school'' is an Elaborator Scene
  in ``The child who went to school is John''.
  A Scene may also be a Participant in another Scene. For example, ``John went to school'' in the sentence: ``He said John went to school''. 
  In other cases, Scenes are annotated as Parallel Scenes (H), which are flat structures and may include a Linker (L), 
  as in: ``When$_L$ [he arrives]$_H$, [he will call them]$_H$''.

  Non-Scene units are headed by units of the category Center,
  denoting the type of entity or thing described by the whole unit.
  Elements in non-Scene units include Quantifiers (such as ``{\it dozens} of people'') and
  Connectors (mostly coordinating conjunctions).
  Other modifiers to the Center are marked as Elaborators.
  
\begin{figure}[th]
  \centering
  \scalebox{.8}{
    \begin{tikzpicture}[level distance=9mm, sibling distance=19mm, ->,
        every circle node/.append style={fill=black}]
      \tikzstyle{word} = [font=\rmfamily,color=black]
      \node (ROOT) [circle] {}
        child {node (After) [word] {After} edge from parent node[above] {\scriptsize L}}
        child {node (graduation) [circle] {}
        {
          child {node [word] {graduation} edge from parent node[left] {\scriptsize P}}
        } edge from parent node[right] {\scriptsize H} }
        child {node [word] {,} edge from parent node[below] {\scriptsize U}}
        child {node (moved) [circle] {}
        {
          child {node (John) [word] {John} edge from parent node[above] {\scriptsize A}}
          child {node [word] {moved} edge from parent node[left] {\scriptsize P}}
          child {node [circle] {}
          {
            child {node [word] {to} edge from parent node[left] {\scriptsize R}}
            child {node [word] {Paris} edge from parent node[right] {\scriptsize C}}
          } edge from parent node[above] {\scriptsize A} }
        } edge from parent node[above] {\scriptsize H} }
        ;
      \draw[dashed,->] (graduation) to node [above] {\scriptsize A} (John);
    \end{tikzpicture}
    }
\caption{UCCA graph. Dashed: remote edge.\label{fig:example_ucca}}
\end{figure}
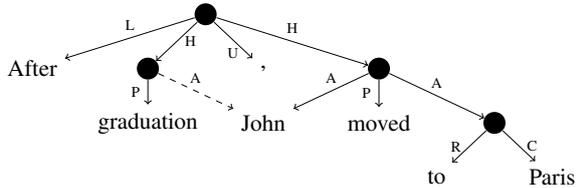

  UCCA distinguishes \textit{primary} edges, corresponding 
  to explicit relations, from \textit{remote} edges,
  which allow for a unit to participate
  in several super-ordinate relations.
  See example in Figure~\ref{fig:example_ucca}.
  Primary edges form a tree, whereas remote edges (dashed) enable reentrancy, forming a DAG.

\paragraph{UD}\label{sec:ud}
is a syntactic dependency scheme used in many languages,
aiming for cross-linguistically consistent and coarse-grained treebank
annotation. Formally, UD uses bi-lexical trees, with edge labels 
representing syntactic relations.

  One aspect of UD similar to UCCA is its preference of lexical (rather than functional) heads.
  For example, in auxiliary verb constructions (e.g., ``is eating''), UD
  marks the lexical verb (\textit{eating}) as the head, while other dependency schemes
  may select the auxiliary \textit{is} instead.
  While the approaches are largely inter-translatable
  \citep{Schwartz:12}, lexical head schemes are more similar in form to semantic schemes,
   such as UCCA and semantic dependencies \citep{oepen2016towards}.
   
Being a dependency representation, UD is structurally underspecified in an important way:
​it is not possible in UD to mark the distinction between an element modifying the head of
the phrase and the same element modifying the whole phrase
\cite{doi:10.1146/annurev-linguistics-011718-011842}.
   
An example UD tree is given in Figure~\ref{fig:original_example_ud}.
UD relations will be written in \texttt{typewriter} font.

\begin{figure}[th]
  \centering
    \begin{dependency}[text only label, label style={above,font=\tt}, font=\small]
    \begin{deptext}[column sep=.8em,ampersand replacement=\^]
    After \^ graduation \^ , \^ John \^ moved \^ to \^ Paris \\
    \end{deptext}
        \depedge[edge unit distance=1ex]{2}{1}{case}
        \depedge[edge unit distance=1ex]{2}{3}{punct}
        \depedge[edge unit distance=1ex]{5}{4}{nsubj}
        \depedge[edge unit distance=1ex, edge end x offset=-2pt]{5}{2}{obl}
        \depedge[edge unit distance=1ex]{7}{6}{case}
        \deproot[edge unit distance=1.5ex]{5}{root}
        \depedge[edge unit distance=1.5ex]{5}{7}{obl}
    \end{dependency}
\caption{UD tree.\label{fig:original_example_ud}}
\end{figure}

\section{Shared Gold-standard Corpus}\label{sec:shared}

We annotate 723 English passages (3,813 sentences; 52,721 tokens),
comprising the web reviews section of the 
English Web Treebank \cite[EWT; ][]{bies2012english}.
Text is annotated by two UCCA annotators
according to v2.0 of the UCCA
guidelines\footnote{\url{http://bit.ly/ucca_guidelines_v2}}
and cross-reviewed.
As these sentences are included in the UD
English\_EWT treebank, this is a \textit{shared} gold-standard UCCA and UD
annotated corpus.\footnote{Our data is available at \url{https://github.com/UniversalConceptualCognitiveAnnotation/UCCA_English-EWT}.}
We use the standard train/development/test split,
shown in Table~\ref{tab:data}.

\begin{table}[t]
\centering
\begin{tabular}{l|ccc}
& \bf \footnotesize Train & \bf \footnotesize Dev & \bf \footnotesize Test \\
\hline
\# Passages & \hphantom{00,}347 & \hphantom{0,}192 & \hphantom{0,}184 \\
\# Sentences & \hphantom{0}2,723 & \hphantom{0,}554 & \hphantom{0,}535 \\
\# Tokens & 44,804 & 5,394 & 5,381 \\
\end{tabular}
\caption{Data split for the shared gold-standard corpus.\label{tab:data}}
\end{table}

\section{Comparison Methodology}\label{sec:methodology}

To facilitate comparison between UCCA and UD,
we first assimilate the graphs by abstracting away from formalism differences,
obtaining a similar graph format for both schemes.
We then match pairs of nodes in the converted UD and UCCA trees
if they share all terminals in their yields.

UD annotates bi-lexical dependency trees,
while UCCA graphs contain non-terminal nodes.
In \S\ref{sec:conversion}, we outline the unified DAG converter by
\citet{hershcovich2018multitask,hershcovich2018universal},\footnote{\url{https://github.com/huji-nlp/semstr}}
which we use to reach a common format.
In \S\ref{sec:local}, we describe a number of extensions
to the converter, which abstract away from further non-content differences.

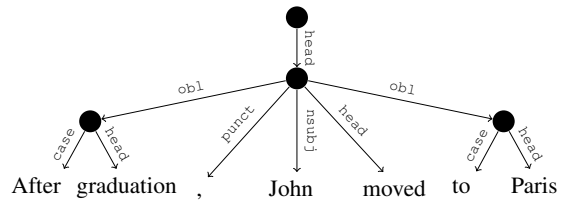
\begin{figure}[th]
  \centering
  \scalebox{.8}{
  \begin{tikzpicture}[level distance=12mm, ->,
      every node/.append style={sloped,anchor=south,auto=false,font=\scriptsize\tt},
      level 2/.style={sibling distance=17mm,level distance=9mm},
      level 3/.style={sibling distance=12mm,level distance=14mm}]
    \tikzstyle{word} = [font=\rmfamily,color=black]
    \node [fill=black,circle] {}
      child {node (ROOT) [fill=black,circle] {}
      {
        child {node (after) [fill=black,circle] {}
        {
          child {node [word] {After{\color{white}g}\quad\quad} edge from parent node {case}}
          child {node [word] {\quad graduation\quad\quad} edge from parent node {head}}
        } edge from parent node {obl}}
        child {node {}
        {
          child {node [word] (comma) {\quad,{\color{white}g}} edge from parent [draw=none]}
        } edge from parent [draw=none]}
        child {node {}
        {
          child {node [word] (John) {John{\color{white}g}} edge from parent [draw=none]}
        } edge from parent [draw=none]}
        child {node {}
        {
          child {node [word] (moved) {moved{\color{white}g}} edge from parent [draw=none]}
        } edge from parent [draw=none]}
        child {node (to) [fill=black,circle] {}
        {
            child {node [word] {to{\color{white}g}} edge from parent node {case}}
            child {node [word] {Paris{\color{white}g}} edge from parent node {head}}
        } edge from parent node {obl}}
      } edge from parent node {head}}
      ;
      \draw (ROOT) to node {punct} (comma);
      \draw (ROOT) to node {nsubj} (John);
      \draw (ROOT) to node {head} (moved);
  \end{tikzpicture}}
\caption{Converted UD tree.
Non-terminals and \textit{head} edges are introduced by the unified DAG converter.\label{fig:converted_example_ud}}
\end{figure}

\subsection{Basic Conversion}\label{sec:conversion}

Figure~\ref{fig:converted_example_ud} presents the same tree from Figure~\ref{fig:original_example_ud}
after conversion.
The converter adds one pre-terminal per token,
and attaches them according to the original dependency tree:
traversing it from the root, for each head it creates a non-terminal
parent with the edge label {\it head}, and adds the dependents as children of 
the created non-terminal.
Relation subtypes are stripped,
leaving only universal relations.
For example, the language-specific definite article label
\texttt{det:def} is replaced by the universal \texttt{det}.

\paragraph{Reentrancies.}
Remote edges in UCCA enable reentrancy, forming a DAG together with primary edges.
UD allows reentrancy when including \textit{enhanced dependencies}
\cite{SCHUSTER16.779},\footnote{\url{https://universaldependencies.org/u/overview/enhanced-syntax.html}} which form (bi-lexical) graphs, representing phenomena
such as predicate ellipsis (e.g., gapping),
and shared arguments due to coordination, control, raising and relative clauses.

UCCA is more inclusive in its use of remote edges, and accounts for 
the entire class of implicit arguments termed {\it Constructional Null Instantiation} in FrameNet \citep{Ruppenhofer:16}.
For example, in
``The Pentagon is bypassing official US intelligence channels [...] in order to create strife'' (from EWT),
remote edges mark \textit{Pentagon} as a shared argument of \textit{bypassing} and
\textit{create}. 
Another example is ``if you call for an appointment [...] so you can then make one'',
where a remote edge in UCCA indicates that \textit{one} refers to \textit{appointment}.
Neither is covered by enhanced UD.

In order to facilitate comparison, we remove remote edges and enhanced dependencies in the conversion process.
We thus compare basic UD and UCCA trees, deferring a comparison of UCCA and enhanced UD to future work.

\subsection{Extensions to the Converter}\label{sec:local}

We extend the unified DAG converter to remove further non-content differences.

\paragraph{Unanalyzable units.}
An unanalyzable phrase is represented in UCCA as a single unit covering multiple terminals.
In multi-word expressions (MWEs) in UD, each word after the first is attached to the previous word,
with the \texttt{flat}, \texttt{fixed} or \texttt{goeswith} relations
(depending on whether the expression is grammaticalized, or split by error).
We remove edges of these relations and join the corresponding pre-terminals to one unit.

\paragraph{Promotion of conjunctions.}
The basic conversion generally preserves terminal yields:
the set of terminals spanned by a non-terminal is the same
as the original dependency yield of its head terminal
(e.g., in Figure~\ref{fig:converted_example_ud}, the yield of the non-terminal
headed by \textit{graduation} is ``After graduation'', the same as that of ``graduation''
in Figure~\ref{fig:original_example_ud}).

Since UD attaches subordinating and coordinating conjunctions to the subsequent conjunct,
this results in them being positioned in the same conjunct they relate (e.g.,
\textit{After} will be included in the first conjunct in ``After arriving home, John went to sleep'';
\textit{and} will be included in the second conjunct in ``John and Mary'').
In contrast, UCCA places conjunctions as siblings to their conjuncts (e.g.,
``[After] [arriving home], [John went to sleep]'' and ``[John] [and] [Mary]''). 

To abstract away from these convention differences,
we place 
coordinating and subordinating conjunctions 
(i.e., \texttt{cc}-labeled units, and \texttt{mark}-labeled units with an \texttt{advcl} head such 
as \textit{when}, \textit{if}, \textit{after}) as siblings of their conjuncts.

\section{Analysis of Divergences}\label{sec:analysis}

Using the shared format,
we turn to analyzing the content differences between UCCA and UD.\footnote{See
\url{http://bit.ly/uccaud} for a detailed explanation of each example in
this section.}

\begin{table*}[t]
\centering
\scalebox{.68}{
\setlength\tabcolsep{6.7pt}
\def\arraystretch{.815}
\begin{tabular}{lRRRRRRRRRRRRRRRR|R}
 &&&&&&&&&&&&&&&&& \multicolumn{1}{|c}{\sc No} \\
 & \multicolumn{1}{c}{\bf A}
 & \multicolumn{1}{c}{\bf A\big|P} & \multicolumn{1}{c}{\bf A\big|S}
 & \multicolumn{1}{c}{\bf C}
 & \multicolumn{1}{c}{\bf D}
 & \multicolumn{1}{c}{\bf E}
 & \multicolumn{1}{c}{\bf F}
 & \multicolumn{1}{c}{\bf G}
 & \multicolumn{1}{c}{\bf H}
 & \multicolumn{1}{c}{\bf L}
 & \multicolumn{1}{c}{\bf N}
 & \multicolumn{1}{c}{\bf P}
 & \multicolumn{1}{c}{\bf Q}
 & \multicolumn{1}{c}{\bf R}
 & \multicolumn{1}{c}{\bf S}
 & \multicolumn{1}{c}{\bf T}
 & \multicolumn{1}{|c}{\sc Match} \\
\tt acl & 58 &  &  & 1 & 4 & 249 & 1 &  & 48 &  &  & 6 &  &  & 1 & 1 & 409 \\
\tt advcl & 14 &  &  & 12 & 2 & 2 &  & 6 & 512 & 4 &  & 11 &  &  &  &  & 423 \\
\tt advmod & 225 &  & 1 & 69 & 1778 & 332 & 27 & 135 & 14 & 258 & 2 & 2 & 15 & 44 & 9 & 368 & 273 \\
\tt amod & 25 &  &  & 134 & 647 & 837 &  & 1 & 28 &  &  & 7 & 130 & 3 & 269 & 25 & 176 \\
\tt appos & 21 &  &  & 39 & 2 & 34 &  &  & 18 &  &  &  &  &  & 8 &  & 33 \\
\tt aux &  &  &  &  & 384 & 2 & 1335 &  &  & 2 &  & 1 &  & 1 &  &  & 17 \\
\tt case & 11 &  &  & 31 & 27 & 25 & 123 &  &  & 213 & 26 & 11 & 1 & 2629 & 154 & 1 & 262 \\
\tt cc &  &  &  & 8 & 4 & 1 & 4 & 1 & 1 & 1567 & 381 &  & 6 & 12 &  &  & 52 \\
\tt ccomp & 345 &  &  & 1 &  & 1 &  &  & 36 &  &  & 2 &  &  & 1 & 1 & 166 \\
\tt compound & 225 &  &  & 116 & 67 & 586 & 21 &  & 2 &  &  & 32 & 19 & 1 & 12 & 24 & 683 \\
\tt conj & 10 &  &  & 449 & 4 & 5 &  & 1 & 1262 & 1 &  & 6 & 2 &  & 10 &  & 497 \\
\tt cop &  &  &  & 1 &  &  & 1312 &  &  & 1 &  & 9 &  & 10 & 178 &  & 7 \\
\tt csubj & 13 &  &  &  &  &  &  &  & 3 &  &  &  &  &  &  &  & 46 \\
\tt det & 10 &  &  & 17 & 119 & 440 & 2963 &  &  &  & 1 &  & 129 & 16 & 1 &  & 124 \\
\tt discourse & 1 &  &  & 2 & 1 &  & 25 & 29 & 27 & 16 &  &  &  &  & 5 &  & 19 \\
\tt expl & 21 &  &  & 1 &  &  & 98 &  &  &  &  &  &  &  & 17 &  & 3 \\
\tt iobj & 131 &  &  & 1 &  &  & 1 &  &  &  &  &  &  &  &  &  & 10 \\
\tt list & 3 &  &  & 7 & 2 & 1 &  &  & 27 &  &  &  &  &  & 1 &  & 6 \\
\tt mark &  &  &  & 9 & 7 & 1 & 531 & 1 &  & 654 &  &  &  & 407 & 1 & 5 & 143 \\
\tt nmod & 844 & 1 & 1 & 20 & 9 & 786 & 8 & 4 & 12 & 1 & 1 & 20 & 2 & 2 & 11 & 27 & 488 \\
\tt nsubj & 4296 & 7 & 21 & 25 & 3 & 2 & 55 & 1 & 5 & 61 &  & 58 & 1 & 80 & 14 & 4 & 247 \\
\tt nummod & 2 &  &  & 33 & 12 & 17 &  & 4 &  & 4 &  &  & 334 &  &  &  & 64 \\
\tt obj & 1845 &  & 1 & 54 & 21 & 6 & 11 & 1 & 4 & 23 &  & 52 & 1 & 23 & 3 & 11 & 583 \\
\tt obl & 1195 &  &  & 19 & 115 & 41 & 1 & 17 & 39 & 34 &  & 6 & 6 & 26 & 7 & 302 & 611 \\
\tt parataxis & 6 &  & 1 & 5 &  & 4 &  & 6 & 285 &  &  &  &  &  & 3 &  & 180 \\
\tt vocative & 17 &  &  &  &  &  &  & 8 &  &  &  &  &  &  &  &  &  \\
\tt xcomp & 121 &  &  & 4 & 25 &  &  &  & 8 &  &  & 38 &  &  & 38 &  & 526 \\
\hline
head & 445 & 48 & 159 & 6388 & 717 & 142 & 564 & 83 & 2462 & 42 & 1 & 4163 & 120 & 52 & 1547 & 32 & 2235 \\
\hline
\sc No Match & 1421 & 37 & 58 & 640 & 417 & 291 & 14 & 33 & 2291 & 146 & 6 & 802 & 94 & 52 & 369 & 96 & 
\end{tabular}
}
\caption{UD-UCCA confusion matrix calculated based on EWT
gold-standard annotations from the training and development sets
(\S\ref{sec:shared}),
after applying our extended converter to UD (\S\ref{sec:methodology}),
by matching UD vertices and UCCA units with the same terminal yield.
The last column (row), labeled {\sc No Match}, shows the number of edges of each UD (UCCA) category
that do not match any UCCA (UD) unit.
Zero counts are omitted.\label{tab:confusion_matrix}}
\end{table*}

\subsection{Confusion Matrix}\label{sec:confusion}

Table~\ref{tab:confusion_matrix} presents the confusion matrix of categories between
the converted UD and UCCA, calculated over all sentences in the training and
development sets of the shared EWT reviews corpus.
We leave the test set out of this evaluation to avoid contamination for future
parsing experiments.

In case of multiple UCCA units with the same terminal yield (i.e., units with a single non-remote child),
we take the top category only, to avoid double-counting.
Excluding punctuation, this results in 60,434 yields in UCCA and
58,992 in UD.
Of these, 52,280 are common, meaning that a UCCA ``parser'' developed this way
would get a very high F1 score
of 87.6\%, if it is provided with the gold UCCA label for every converted edge.

Some yields still have more than one UCCA category associated with them,
due to edges with multiple categories ({\bf A\big|P} and {\bf A\big|S}).
For presentation reasons, 0.15\% of the UCCA units in the data
are not presented here, as they belong to rare (<~0.1\%)
multiple-category combinations.

Only 82.6\% of UD's syntactic arguments
(\texttt{ccomp}, \texttt{csubj}, \texttt{iobj}, \texttt{nsubj}, \texttt{obj}, \texttt{obl} and \texttt{xcomp})
are UCCA Participants,
and only 72.9\% of the Participants are syntactic arguments---a difference stemming from
the Scene/non-Scene (\S\ref{sec:s}) and argument/adjunct (\S\ref{sec:arguments}) distinctions.
Moreover, if we identify predicates as words having at least one argument
and Scenes as units with at least one Participant,
then only 92.1\% of UD's predicates correspond to Scenes (many are secondary relations within one scene),
and only 80\% of Scenes correspond to predicates (e.g., eventive nouns,
which are not syntactic predicates).

Examining the {\it head} row in Table~\ref{tab:confusion_matrix} allows
us to contrast the schemes' notions of a head. 
{\it head}-labeled units have at least
one dependent in UD, or are single-clause sentences (technically, they are non-terminals added by the converter).
Of them, 75.7\% correspond to Processes, States, Parallel Scenes or Centers,
which are UCCA's notions of semantic heads,
and 11.6\% are left unmatched, mostly due to MWEs analyzed in
UD but not in UCCA (\S\ref{sec:mwe}).
Another source of unmatched units is inter-Scene linkage, which tends to be flatter in
UCCA (\S\ref{sec:linkage}).
The rest are mostly due to head swap (e.g., ``\textit{all} of Dallas'', where \textit{all}
is a Quantifier of \textit{Dallas} in UCCA, but the head in UD).

In the following subsections, we review the main content differences between the schemes,
as reflected in the confusion matrix, and categorize them according to the UD relations
involved.

\subsection{Scenes vs. Non-Scenes}\label{sec:s}

UCCA distinguishes between Scenes and non-Scenes.
This distinction crosses UD categories,
as a Scene can be evoked by a verb, an eventive or stative
noun (\textit{negotiation}, \textit{fatigue}),
an adjective or even a preposition (``this is \textit{for} John'').

\paragraph{Core syntactic arguments.}
      Subjects and objects are usually Participants (e.g., ``\textit{wine} was excellent'').
      However, when describing a Scene, the subject may be a Process/State
      (e.g., ``but \textit{service} is very poor'').
      Some wh-pronouns are the subjects or objects of a relative clause, but
      are Linkers or Relators,
      depending on whether they link Scenes or non-Scenes, respectively.
      For example, ``who'' in ``overall, Joe is a happy camper \textit{who} has found a great spot'' is an \texttt{nsubj}, but a Linker.
      Other arguments are Adverbials or Time (see \S\ref{sec:arguments}), and
      some do not match any UCCA unit, especially when they are parts of MWEs (see \S\ref{sec:mwe}).

\paragraph{Adjectival modifiers} are Adverbials when modifying Scenes
    (``\textit{romantic} dinner''), States when describing non-Scenes (``\textit{beautiful} hotel'') 
    or when semantically predicative (``such a \textit{convenient} location''), or
    Elaborators where defining inherent properties of non-Scenes (``\textit{medical} school'').

\paragraph{Nominal and clausal modifiers.}
    Most are Participants or Elaborators,
    depending on whether they modify a Scene (e.g.,
    ``discount \textit{on services}'' and
    ``our decision \textit{to buy when we did}'' are Participants,
    but ``\textit{my car's} gears and brakes'' and ``Some of the younger kids \textit{that work there}'' are Elaborators).
    Unmatched \texttt{acl} are often
    free relative clauses (e.g., in ``the prices were worth what \textit{I got}'',
    \textit{what} is the \texttt{obj} of \textit{worth} but
    a Participant of \textit{I got}).

\paragraph{Case markers.}
      While mostly Relators
      modifying non-Scenes (e.g., ``the team \textit{at} Bradley Chevron''),
      some case markers are Linkers linking Scenes together 
      (e.g., ``very informative website \textit{with} a lot of good work'').
      Others are Elaborators (e.g., ``\textit{over} a year'') or States
      when used as the main relation in verbless or copula clauses
      (e.g., ``it is right \textit{on} Wisconsin Ave'').
    
\paragraph{Coordination.}
      Coordinating conjunctions (\texttt{cc}) are Connectors where they coordinate non-Scenes
      (e.g., ``Mercedes \textit{and} Dan'')
      or Linkers where they coordinate Scenes (e.g., ``outdated \textit{but} not bad'').
      Similarly, conjuncts and list elements (\texttt{conj}, \texttt{list}) may be Parallel Scenes (H),
      or Centers when they are non-Scenes.\footnote{While in UD 
      the conjunction \texttt{cc} is attached to the following conjunct,
      in UCCA coordination is a flat structure.
      This is a convention difference that we normalize (\S\ref{sec:local}).}

\paragraph{Determiners.}
      Articles are Functions,
      but determiners modifying non-Scenes are Elaborators
      (e.g., ``I will never recommend this gym to \textit{any} woman'').
      Where modifying Scenes (mostly negation)
      they are marked as Adverbials. For example, ``\textit{no} feathers in stock'', ``\textit{what} a mistake'',
      and ``the rear window had \textit{some} leakage'' are all Adverbials.

\subsection{Primary and Secondary Relations}\label{sec:arguments}

UD distinguishes core arguments, adverb modifiers,
and obliques (in English UD, the latter mostly correspond to prepositional dependents of verbs).
UCCA distinguishes Participants, including locations and abstract entities,
from secondary relations (Adverbials), 
which cover manner, aspect and modality.
Adverbials can be verbs (e.g., \textit{begin}, \textit{fail}),
prepositional phrases (\textit{with disrespect}),
as well as modals, adjectives and adverbs.

\paragraph{Adverbs and obliques.}
    Most UD adverb modifiers are Adverbials (e.g., ``I \textit{sometimes} go''),
    but they may be Participants, mostly in the case of semantic arguments describing location (e.g., \textit{here}).
    Obliques
    may be
    Participants (e.g., ``wait \textit{for Nick}''), Time (e.g., ``for \textit{over 7 years}'') 
    or Adverbials---mostly manner adjuncts (\textit{by far}).

\paragraph{Clausal arguments}
    are Participant Scenes
    (e.g., ``it was great \textit{that they did not charge a service fee}'',
    ``did not really know \textit{what I wanted}'' or
    ``I asked them \textit{to change it}'').
    However, when serving as complements to a secondary verb, they
    will not match any unit in UCCA, as it places secondary verbs on the 
    same level as their primary relation. 
    For example, \textit{to pay} is an \texttt{xcomp} in ``they have to pay'', while
    the UCCA structure is flat: \textit{have to} is an Adverbial and \textit{pay} is a Process.
    Single-worded clausal arguments may correspond to a Process/State,
    as in ``this seems \textit{great}''.

\paragraph{Auxiliary verbs}
    are Functions (e.g., ``\textit{do} not forget''),
    or Adverbials when they are modals (e.g., ``you \textit{can} graduate''). Semi-modals 
    in UD are treated as clausal heads, which take a clausal complement. 
    For example, in ``able to do well'', UD treats \textit{able} as the head,
    which takes \textit{do well} as an \texttt{xcomp}. UCCA, on the other hand,
    treats it as an Adverbial, creating a mismatch for \texttt{xcomp}.

\subsection{Multi-Word Expressions}\label{sec:mwe}

UD and UCCA treat MWEs differently.
In UD they include names, compounds and grammaticalized fixed expressions.
UCCA treats names and grammaticalized MWEs as unanalyzable units,
but also a range of semantically opaque constructions
(e.g., light verbs and idioms).
On the other hand, compounds are not necessarily unanalyzable in UCCA,
especially if compositional.

\paragraph{Compounds.} English compounds are mostly nominal,
        and are a very heterogeneous category.
        Most compounds correspond to Elaborators (e.g., ``\textit{industry} standard''),
        or Elaborator Scenes (e.g., ``\textit{out-of-place} flat-screen TV''),
        and many are unanalyzable expressions (e.g., ``\textit{mark} up'').
        Where the head noun evokes a Scene, the dependent is often a Participant
        (e.g., ``\textit{food} craving''), but can also be an Adverbial 
        (e.g., ``\textit{first time} buyers'') depending on its semantic category.
        Other compounds in UD are phrasal verbs (e.g., ``figure \textit{out}'', ``cleaned \textit{up}''),
        which UCCA treats as unanalyzable (leading to unmatched units). 
            
\paragraph{Core arguments.}
      A significant number of subjects and objects are left unmatched as they
      form parts of MWEs marked in UCCA as unanalyzable. UD annotates
      MWEs involving a verb and its argument(s) just like any other clause, and therefore
      lacks this semantic content. Examples include light verbs (e.g., ``give {\it a try}''),
      idioms (``bites {\it the dust}''), and figures of speech (e.g., ``when \textit{it} comes to'', ``offer \textit{a taste} (of)''),
      all are UCCA units.
      
\paragraph{Complex prepositions.} Some complex prepositions (e.g., \textit{according to} or \textit{on top of}),
      not encoded as MWEs in UD, are unanalyzable in UCCA.

\subsection{Linkage}\label{sec:linkage}

\begin{figure}[th]
  \centering
\begin{subfigure}{.45\columnwidth}
    \begin{dependency}[text only label, label style={above,font=\tt}, font=\small, edge unit distance=1.5ex]
    \begin{deptext}[column sep=.1em,ampersand replacement=\^]
    unique \^ gifts \^ and \^ cards \\
    \end{deptext}
        \depedge{2}{1}{amod}
        \deproot{2}{root}
        \depedge{4}{3}{cc}
        \depedge{2}{4}{conj}
    \end{dependency}
    \caption{UD\label{fig:conj_ud}}
\end{subfigure}
\hfill
\begin{subfigure}{.45\columnwidth}
    \scalebox{.8}{
    \begin{tikzpicture}[level distance=9mm, sibling distance=14mm, ->,
        every circle node/.append style={fill=black}]
      \tikzstyle{word} = [font=\rmfamily,color=black]
      \node (ROOT) [circle] {}
        child {node [word] {unique} edge from parent node[left] {\scriptsize E}}
        child {node [circle] {}
        {
          child {node [word] {gifts} edge from parent node[left] {\scriptsize C}}
          child {node [word] {and} edge from parent node[left] {\scriptsize N}}
          child {node [word] {cards} edge from parent node[right] {\scriptsize C}}
        } edge from parent node[right] {\scriptsize C} }
        ;
    \end{tikzpicture}
    }
    \caption{UCCA\label{fig:conj_ucca}}
 \end{subfigure}
 \caption{Coordination in UD and UCCA.\label{fig:conj}}
\end{figure}
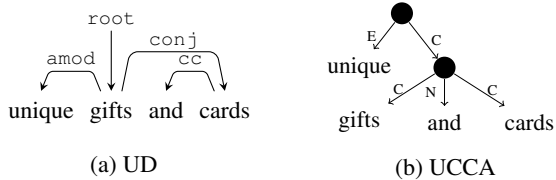

\paragraph{Head selection.}
UCCA tends to flatten linkage, where UD, as a dependency scheme,
selects a head and dependent per relation.
This yields scope ambiguities for coordination, an inherently flat structure. 
For instance, ``unique gifts and cards'' is ambiguous in UD as to whether
\textit{unique} applies only to \textit{gifts} or to the whole phrase---both
annotated as in Figure~\ref{fig:conj_ud}.
UCCA, allowing non-terminal nodes, disambiguates this case (Figure~\ref{fig:conj_ucca}).

\paragraph{Clausal dependents.}
UD categorizes clause linkage into coordination,
subordination, argumenthood (complementation),
and parataxis. 
UCCA distinguishes argumenthood 
but conflates the others into the Parallel Scene category.
For example,
``We called few companies before \textit{we decided to hire them}''
and ``Check out The Willow Lounge, \textit{you'll be happy}'' are Parallel Scenes.

\begin{figure}[th]
\begin{subfigure}{\columnwidth}
  \begin{minipage}[c]{.12\textwidth}
    \caption{UD\label{fig:linkage_ud}}
  \end{minipage}
  \begin{minipage}[c]{.8\textwidth}
    \begin{dependency}[text only label, label style={above,font=\tt}, font=\small, edge unit distance=1.8ex]
    \begin{deptext}[column sep=.25em,ampersand replacement=\^]
    From \^ the \^ moment \^ you \^ enter \^ , \^ you \^ know \\
    \end{deptext}
        \depedge[edge unit distance=2.4ex]{3}{1}{case}
        \depedge{3}{2}{det}
        \depedge[edge unit distance=1ex]{8}{3}{obl}
        \depedge[edge unit distance=1.5ex]{3}{5}{acl}
        \depedge[edge start x offset=-5pt,edge unit distance=1ex]{5}{4}{nsubj}
        \depedge[edge unit distance=1.5ex]{8}{6}{punct}
        \depedge[edge start x offset=-10pt,edge unit distance=1ex]{8}{7}{nsubj}
        \deproot{8}{root}
    \end{dependency}
  \end{minipage}
\end{subfigure}

\begin{subfigure}{\columnwidth}
  \begin{minipage}[c]{.17\textwidth}
    \caption{UCCA\label{fig:linkage_ucca}}
  \end{minipage}
  \hspace{-8mm}
  \begin{minipage}[c]{.8\textwidth}
    \scalebox{.8}{
    \begin{tikzpicture}[level distance=9mm, sibling distance=28mm, ->,
        level 2/.style={sibling distance=9mm},
        every circle node/.append style={fill=black}]
      \tikzstyle{word} = [font=\rmfamily,color=black]
      \node (ROOT) [circle] {}
        child {node [circle] {}
        {
          child {node [word] {From} edge from parent node[left] {\scriptsize R}}
          child {node [word] {the} edge from parent node[left] {\scriptsize E}}
          child {node [word] {{\color{white}F}moment} edge from parent node[right] {\scriptsize C}}
        } edge from parent node[above] {\scriptsize L} }
        child {node [circle] {}
        {
          child {node [word] {you{\color{white}k}} edge from parent node[left] {\scriptsize A}}
          child {node [word] {{\color{white}y}enter} edge from parent node[right] {\scriptsize P}}
        } edge from parent node[left] {\scriptsize H} }
        child {node [circle] {}
        {
          child {node (comma) [word] {,} edge from parent [draw=none]}
          child {node [word] {you{\color{white}k}} edge from parent node[left] {\scriptsize A}}
          child {node [word] {{\color{white}y}know} edge from parent node[right] {\scriptsize S}}
        } edge from parent node[above] {\scriptsize H} }
        ;
      \draw[->] (ROOT) to node [right] {\scriptsize U} (comma);
    \end{tikzpicture}
    }
  \end{minipage}
 \end{subfigure}
 \caption{Clause linkage in UD and UCCA.\label{fig:linkage}}
\end{figure}
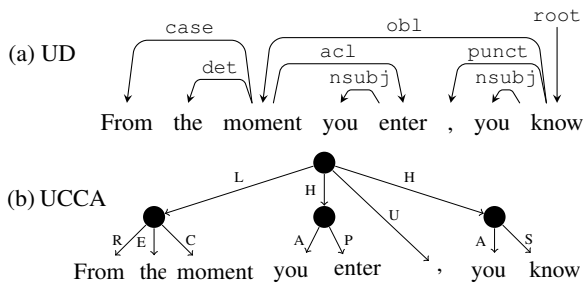

Note that while in UD, \texttt{mark} (e.g., \textit{before})
is attached to the dependent adverbial clause,
a UCCA Linker lies outside the linked Scenes.
To reduce unmatched \texttt{advcl} instances,
this convention difference is fixed by the converter
(\S\ref{sec:local}).
Many remaining unmatched units are due to conjunctions we could not reliably raise.
For instance, the marker \textit{to} introducing an \texttt{xcomp} is ambiguous between Linker
(purposive \textit{to}) and Function (infinitive marker).
Similarly, wh-pronouns may be Linkers
(``he was willing to budge a little on the price {\it which} means a lot to me''),
but have other uses in questions and free relative clauses.
Other mismatches result from the long tail of differences in how UD and UCCA construe linkage.
Consider the sentence in Figure~\ref{fig:linkage}.
While \textit{moment} is an oblique argument of \textit{know} in UD,
\textit{From the moment} is analyzed as a Linker in UCCA.

\subsection{Other Differences}\label{sec:misc}

\paragraph{Appositions}
    in UD always follow the modified noun,
    but named entities in them are UCCA Centers, regardless of position
    (e.g., in ``its sister store Peking Garden'', the UD head \textit{its sister store}
    is an Elaborator, while \textit{Peking Garden} is the Center).

\paragraph{Copulas.}
    UCCA distinguishes copular constructions expressing
    identity (e.g., ``This \textit{is} the original Ham's restaurant'') where the copula is annotated as State,
    and cases of attribution 
    (e.g., ``Mercedes and Dan \textit{are} very thorough'')
    or location (e.g., ``Excellent chefs \textit{are} in the kitchen''),
    where the copula is a Function.

\paragraph{Discourse markers and interjections.}
    Units relating a Scene to the speech event or to the speaker's opinion are Ground
    (e.g., ``\textit{no}, Warwick in New Jersey'' and ``\textit{Please} visit my website'').
    On the other hand, discourse elements that relate one Scene to another 
    are Linkers (e.g., \textit{anyway}).

\paragraph{Vocatives}
    are both Ground and Participants if they participate in the Scene \textit{and} are the party addressed.
    For example, \textit{Mark} in ``Thanks \textit{Mark}'' is both the person addressed and the one thanked.\footnote{The {\bf A\big|G} column is omitted from
    Table~\ref{tab:confusion_matrix} as this category combination
    occurs in only 0.02\% of edges in the corpus.}
    
\paragraph{Expletives and subjects.}
    Expletives are generally Functions,
    but some instances of \textit{it} and \textit{that} are analyzed as \texttt{nsubj} in UD
    and as Function in UCCA (e.g., ``\textit{it}'s like driving a new car'').

\paragraph{Excluded relations.}
We exclude the following UD labels,
as they are irrelevant to our evaluation:
\texttt{root} (always matches the entire sentence);
\texttt{punct} (punctuation is ignored in UCCA evaluation);
\texttt{dep} (unspecified dependency),
\texttt{orphan} (used for gapping, which is represented using remote edges in UCCA---see \S\ref{sec:conversion});
\texttt{fixed}, \texttt{flat} and \texttt{goeswith} (correspond to parts of unanalyzable units in UCCA,
    and so do not represent units on their own---see \S\ref{sec:local});
    \texttt{reparandum} and \texttt{dislocated} (too rare in EWT).

\section{Fine-Grained UCCA Parsing Evaluation}\label{sec:fine_grained}

In \S\ref{sec:analysis} we used our comparison methodology,
consisting of the conversion to a shared format and matching units by terminal yield,
to compare gold-standard UD and UCCA.
In this section we apply the same methodology to parser outputs,
using gold-standard UD for fine-grained evaluation.

\subsection{Experimental Setup}\label{sec:experiments}

\paragraph{Data.}

In addition to the UCCA EWT data (\S\ref{sec:shared}),
we use the reviews section of the UD v2.3 English\_EWT treebank
\cite{11234/1-2895},\footnote{\url{https://hdl.handle.net/11234/1-2895}}
annotated over the exact same sentences.
We additionally use UDPipe v1.2 \cite{udpipe,udpipe:2017},
trained on
English\_EWT,\footnote{\url{https://hdl.handle.net/11234/1-2898}}
for feature extraction.
We apply the extended converter to UD as before (\S\ref{sec:local}).

\paragraph{Parser.}

We train TUPA v1.3 \cite{hershcovich2017a,hershcovich2018multitask}
on the UCCA EWT data, with the standard train/development/test split.
TUPA uses POS tags and syntactic dependencies as features.
We experiment both with using gold UD for feature extraction,
and with using UDPipe outputs.

\paragraph{Evaluation by gold-standard UD.}

UCCA evaluation is generally carried out by considering a predicted unit as correct if there
is a gold unit that matches it in terminal yield and labels. Precision, Recall and F-score (F1)
are computed accordingly.
For the fine-grained analysis, we split the gold-standard, predicted and matched UCCA units according
to the labels of the UD relations whose dependents have the same terminal yield (if any).

\subsection{Results}\label{sec:results}

Table~\ref{tab:results} presents TUPA's scores on the UCCA EWT development
and test sets. Surprisingly, using UDPipe for feature extraction results in
better scores than gold syntactic tags and dependencies.

\begin{table}[h]
\def\arraystretch{.89}
\resizebox{\columnwidth}{!}{
\begin{tabular}{@{}llll|lll@{}}
& \multicolumn{3}{c|}{\footnotesize \bf Primary} & \multicolumn{3}{c}{\footnotesize \bf Remote} \\
\multicolumn{1}{l}{\footnotesize \bf Features}
& \footnotesize \textbf{LP} & \footnotesize \textbf{LR} & \footnotesize \textbf{LF}
& \footnotesize \textbf{LP} & \footnotesize \textbf{LR} & \footnotesize \textbf{LF} \\
\hline
\multicolumn{4}{@{}l|}{\footnotesize Development} \\
\multicolumn{1}{l}{\footnotesize Gold UD}
 & 72.1 & 71.2 & 71.7 & 61.2 & 38.1 & 47.0 \\
\multicolumn{1}{l}{\footnotesize UDPipe} 
 & 73.0 & 72.1 & 72.5 & 53.7 & 40.8 & 46.4 \\
\hline
\multicolumn{4}{@{}l|}{\footnotesize Test} \\
\multicolumn{1}{l}{\footnotesize Gold UD}
 & 72.2 & 71.2 & 71.7 & 60.9 & 36.8 & 45.9 \\
\multicolumn{1}{l}{\footnotesize UDPipe} 
 & 72.4 & 71.7 & 72.1 & 60.3 & 38.5 & 47.0
\end{tabular}}
\caption{
Labeled precision, recall and F1 (in~\%) for primary and
remote edges output by TUPA on the UCCA EWT development (top) and test (bottom)
sets, using either gold-standard UD or UDPipe for TUPA features.\label{tab:results}}
\end{table}

\begin{table*}[t]
\vspace{-1mm}
\centering
\scriptsize
\setlength\tabcolsep{1.9pt}
\def\arraystretch{1.2}
\hspace{-2mm}
\begin{tabular}{cl|rrrrrrrrrrrrrrrrrrrrrrrrrrr}
 &  & \multicolumn{1}{c}{\scriptsize \tt \rotatebox{90}{aux}} & \multicolumn{1}{c}{\scriptsize \tt \rotatebox{90}{det}} & \multicolumn{1}{c}{\scriptsize \tt \rotatebox{90}{cop}} & \multicolumn{1}{c}{\scriptsize \tt \rotatebox{90}{cc}} & \multicolumn{1}{c}{\scriptsize \tt \rotatebox{90}{expl}} & \multicolumn{1}{c}{\scriptsize \tt \rotatebox{90}{iobj}} & \multicolumn{1}{c}{\scriptsize \tt \rotatebox{90}{nsubj}} & \multicolumn{1}{c}{\scriptsize \tt \rotatebox{90}{case}} & \multicolumn{1}{c}{\scriptsize \tt \rotatebox{90}{list}} & \multicolumn{1}{c}{\scriptsize \tt \rotatebox{90}{advmod}} & \multicolumn{1}{c}{\scriptsize \tt \rotatebox{90}{amod}} & \multicolumn{1}{c}{\scriptsize \tt \rotatebox{90}{nummod}} & \multicolumn{1}{c}{\scriptsize \tt \rotatebox{90}{mark}} & \multicolumn{1}{c}{\scriptsize \tt \rotatebox{90}{vocative}} & \multicolumn{1}{c}{\scriptsize \tt \rotatebox{90}{compound}} & \multicolumn{1}{c}{\scriptsize \tt \rotatebox{90}{obj}} & \multicolumn{1}{c}{\scriptsize \tt \rotatebox{90}{nmod}} & \multicolumn{1}{c}{\scriptsize \tt \rotatebox{90}{conj}} & \multicolumn{1}{c}{\scriptsize \tt \rotatebox{90}{advcl}} & \multicolumn{1}{c}{\scriptsize \tt \rotatebox{90}{obl}} & \multicolumn{1}{c}{\scriptsize \tt \rotatebox{90}{xcomp}} & \multicolumn{1}{c}{\scriptsize \tt \rotatebox{90}{discourse}} & \multicolumn{1}{c}{\scriptsize \tt \rotatebox{90}{ccomp}} & \multicolumn{1}{c}{\scriptsize \tt \rotatebox{90}{parataxis}} & \multicolumn{1}{c}{\scriptsize \tt \rotatebox{90}{appos}} & \multicolumn{1}{c}{\scriptsize \tt \rotatebox{90}{acl}} & \multicolumn{1}{c}{\scriptsize \tt \rotatebox{90}{csubj}} \\ \hline
\multirow{2}{*}{\footnotesize (a)} & \scriptsize \bf Labeled F1 \% & 94 & 93 & 89 & 86 & 83 & 83 & 80 & 76 & 76 & 72 & 71 & 71 & 70 & 62 & 59 & 57 & 55 & 50 & 49 & 48 & 41 & 38 & 29 & 23 & 21 & 20 & 0 \\
 & \scriptsize \bf Unlabeled F1 \% & 99 & 99 & 100 & 99 & 100 & 83 & 84 & 95 & 76 & 95 & 95 & 86 & 97 & 92 & 84 & 65 & 77 & 61 & 51 & 61 & 63 & 95 & 29 & 36 & 48 & 37 & 33 \\ \hline
\multirow{5}{*}{\footnotesize (b)} & \scriptsize \bf Total in UD \# & 156 & 392 & 187 & 212 & 12 & 8 & 463 & 335 & 15 & 378 & 374 & 38 & 116 & 1 & 219 & 222 & 231 & 244 & 52 & 208 & 1 & 16 & 29 & 52 & 22 & 81 & 5 \\
 & \scriptsize \bf Match Gold \# & 156 & 385 & 187 & 206 & 12 & 6 & 468 & 305 & 12 & 359 & 361 & 33 & 111 & 7 & 146 & 187 & 198 & 210 & 40 & 162 & 28 & 10 & 20 & 48 & 17 & 56 & 4 \\
 & \scriptsize \bf Match Predicted \# & 154 & 388 & 187 & 203 & 12 & 6 & 446 & 313 & 9 & 345 & 339 & 32 & 113 & 6 & 136 & 163 & 183 & 177 & 30 & 147 & 26 & 11 & 15 & 30 & 12 & 36 & 2 \\
 & \scriptsize \bf Labeled Correct \# & 145 & 361 & 166 & 175 & 10 & 5 & 365 & 236 & 8 & 253 & 248 & 23 & 78 & 4 & 83 & 99 & 104 & 96 & 17 & 74 & 11 & 4 & 5 & 9 & 3 & 9 & 0 \\
 & \scriptsize \bf Unlabeled Correct \# & 154 & 381 & 187 & 203 & 12 & 5 & 386 & 293 & 8 & 336 & 334 & 28 & 109 & 6 & 118 & 113 & 147 & 119 & 18 & 94 & 17 & 10 & 5 & 14 & 7 & 17 & 1 \\ \hline
\multirow{2}{*}{\footnotesize (c)} & \scriptsize \bf Labeled/Unlabeled \% & 94 & 95 & 89 & 86 & 83 & 100 & 95 & 81 & 100 & 75 & 74 & 82 & 72 & 67 & 70 & 88 & 71 & 81 & 94 & 79 & 65 & 40 & 100 & 64 & 43 & 53 & 0 \\
 & \scriptsize \bf Mode/Match Gold \% & 79 & 82 & 86 & 75 & 58 & 100 & 91 & 79 & 83 & 51 & 35 & 85 & 45 & 71 & 54 & 91 & 51 & 70 & 92 & 68 & 44 & 30 & 94 & 98 & 41 & 72 & 100 \\ \hline
\footnotesize (d) & \scriptsize \bf Average Words \# & 1.0 & 1.0 & 1.0 & 1.0 & 1.0 & 1.1 & 1.6 & 1.0 & 2.2 & 1.2 & 1.2 & 1.1 & 1.0 & 1.6 & 1.2 & 3.0 & 2.4 & 5.8 & 6.6 & 3.8 & 6.0 & 1.1 & 9.0 & 6.7 & 4.0 & 5.6 & 7.5
\end{tabular}
\caption{Fine-grained evaluation of TUPA (with gold-standard UD features) on the
EWT development set.
(a) Columns are sorted by labeled F1, measuring performance on each subset of edges.
Unlabeled F1 ignores edge categories, evaluating unit boundaries only.
(b) Total number of instances of each UD relation;
of them, matching UCCA units in gold-standard and in TUPA's predictions;
their intersection, with/without regard to categories.
(c) 
Percentage of correctly categorized edges;
for comparison, percentage of most frequent category (see~Table~\ref{tab:confusion_matrix}).
(d) Average number of words in corresponding terminal yields.\label{tab:fine_grained_results}}
\end{table*}

Table~\ref{tab:fine_grained_results} shows
fine-grained evaluation by UD relations.
TUPA does best on auxiliaries and determiners,
despite the heterogeneity of corresponding UCCA categories
(see Table~\ref{tab:confusion_matrix}),
possibly by making lexical distinctions
(e.g., modals and auxiliary verbs are both UD auxiliaries,
but are annotated as Adverbials and Functions, respectively).

Copulas and coordinating conjunctions pose a more difficult distinction,
since the same lexical items may have different categories depending on the
context: State/Function for copulas,
due to the distinction between identity and attribution, and
Connector/Linker for conjunctions,
due to the distinction between Scenes and non-Scenes.
However, the reviews domain imposes a strong prior for both (Function and Linker,
respectively), which TUPA learns successfully.

Inter-clause linkage (\texttt{conj}, \texttt{advcl}, \texttt{xcomp},
\texttt{ccomp}, \texttt{parataxis}, \texttt{acl} and \texttt{csubj})
is a common source of error for TUPA.
Although the match between UCCA and UD is not perfect in these cases,
it is overall better than TUPA's unlabeled performance,
despite using gold-standard syntactic features.
Our results thus suggest that encoding syntax more directly, perhaps using syntactic
scaffolding \citep{swayamdipta2018syntactic}
or guided attention \citep{strubell2018linguistically},
may assist in predicting unit boundaries.
However, TUPA often succeeds at making distinctions that are not even encoded in UD.
For example, it does reasonably well (71\%) on distinguishing between noun modifiers of
Scene-evoking nouns (Participants) and modifiers of other nouns (Elaborators),
surpassing a majority baseline based on the UD relation (51\%).
Lexical resources that distinguish eventive and relational nouns from concrete 
nouns may allow improving it even further.
In the similar case of compounds, lexical resources for light verbs and idioms may increase performance.

\section{Discussion}\label{sec:discussion}

NLP tasks
often require semantic distinctions that are difficult to extract from syntactic representations.
Consider the example
``after graduation, John moved to Paris'' again.
While \textit{graduation} evokes a Scene
(Figure~\ref{fig:example_ucca}), in UD it is an oblique modifier of \textit{moved},
just like \textit{Paris} is (Figure~\ref{fig:original_example_ud}).
The Scene/non-Scene distinction (\S\ref{sec:s})
would assist structural text simplification systems
in paraphrasing this sentence to two sentences, each one
containing one Scene \cite{sulem2018samsa}.

Another example is machine translation---translating the same sentence into Hebrew,
which does not have a word for \textit{graduation},
would require a clause to convey the same meaning.
The mapping would therefore be more direct using a semantic representation,
and we would benefit from breaking the utterance into two Scenes.

\section{Related Work}\label{sec:related_work}

The use of syntactic parsing as a proxy for semantic structure has a long tradition in NLP.
Indeed, semantic parsers have leveraged syntax
for output space pruning \cite{xue2004calibrating}, 
syntactic features \cite{gildea2002automatic,hershcovich2017a}, 
joint modeling \cite{surdeanu2008conll,hajivc2009conll}, and
multi-task learning \cite{swayamdipta2016greedy,swayamdipta2018syntactic,hershcovich2018multitask}.
Empirical comparison between syntactic and semantic schemes, however,
is still scarce.
\newcite{W14-2908} mapped Stanford Dependencies
(precursor to UD) to Hobbsian Logical Form, identifying semantic gaps in the former.
PredPatt \citep{white2016universal},
a framework for extracting predicate-argument structures from UD,
was evaluated by \newcite{zhang2017evaluation}
on a large set of converted PropBank annotations.
\newcite{szubert2018structured} proposed a method for aligning AMR and UD subgraphs,
finding that 97\% of AMR edges are evoked by one or more
words or syntactic relations.
\newcite{damonte-17} refined AMR evaluation by UD labels,
similar to our fine-grained evaluation of UCCA parsing.

Some syntactic representation approaches, notably CCG \cite{Steedman:00},
directly reflect the underlying semantics, and have been used to
transduce semantic forms using rule-based systems \cite{Basile:12}.
A related line of work tackles the transduction of syntactic structures into semantic ones.
\newcite{reddy2016transforming} proposed a rule-based method for converting UD
to logical forms.
\newcite{stanovsky2016getting} converted Stanford dependency trees into
proposition structures ({\sc PropS}), abstracting away from some syntactic detail.

\section{Conclusion}\label{sec:conclusion}

We evaluated the similarities and divergences in the content encoded by UD and UCCA. 
We annotated the reviews section of the English Web Treebank with UCCA,
  and used an automated methodology to evaluate how well the two schemes align,
  abstracting away from differences of mere convention.
We provided a detailed picture of the content differences between the schemes.
Notably, we quantified the differences between the notions of syntactic and semantic heads
  and arguments, finding substantial divergence between them.
Our findings highlight the potential utility of using semantic parsers for text understanding applications
  (over their syntactic counterparts), but also expose challenges semantic parsers must address,
  and potential approaches for addressing them.

\section*{Acknowledgments}

This work was supported by the Israel Science Foundation (grant No. 929/17),
and by the HUJI Cyber Security Research Center
in conjunction with the Israel National Cyber Bureau in the Prime Minister's Office.
We thank Jakob Prange, Nathan Schneider
and the anonymous reviewers for their helpful comments.

\bibliography{references}
\bibliographystyle{acl_natbib}

\end{document}